# Empowering Clinical Trial Design through AI: A Randomized Evaluation of PowerGPT

**Authors:**

Yiwen Lu[1,2], Lu Li[1,2], Dazheng Zhang[1,3], Xinyao Jian[1,3], Tingyin Wang[1,2], Siqi Chen[1,2], Yuqing Lei[1,3], Jiayi Tong[1,3,4], Zhaohan Xi[5], Haitao Chu[6,7], Chongliang Luo[8,9], Alexis Ogdie[10], Brian Athey[11], Alparslan Turan[12], Michael Abramoff[13,14], Joseph C Cappelleri[15], Hua Xu[16], Yun Lu[17], Jesse Berlin[18,19,*], Daniel I. Sessler[12,*], David A. Asch[20,21,*], Xiaoqian Jiang[22,*], Yong Chen[1-3,20,23,24,*]

**Author Affiliations:**

1. Center for Health AI and Synthesis of Evidence (CHASE), Department of Biostatistics, Epidemiology and Informatics, Perelman School of Medicine, University of Pennsylvania, Philadelphia, PA, USA
2. The Graduate Group in Applied Mathematics and Computational Science, School of Arts and Sciences, University of Pennsylvania, Philadelphia, PA, USA
3. Department of Biostatistics, Epidemiology, and Informatics, University of Pennsylvania Perelman School of Medicine
4. Department of Biostatistics, Johns Hopkins Bloomberg School of Public Health, Baltimore, MD, USA
5. School of Computing, Binghamton University, The State University of New York (SUNY), Binghamton, NY, USA
6. Division of Biostatistics and Health Data Science, University of Minnesota, Minneapolis, MN, USA
7. Statistical Research and Data Science Center, Pfizer Inc, New York, NY, USA
8. Division of Public Health Sciences, Department of Surgery, Washington University School of Medicine, St. Louis, MO USA
9. Siteman Cancer Center Biostatistics Shared Resource, Division of Public Health Sciences, Department of Surgery, Washington University School of Medicine, St. Louis, MO USA
10. Department of Medicine and Rheumatology, University of Pennsylvania Perelman School of Medicine, Philadelphia, PA, USA
11. Department of Computational Medicine and Bioinformatics, University of Michigan Medical School, Ann Arbor, MI, USA
12. Department of Anesthesiology, Critical Care and Pain Medicine, McGovern Medical School, Houston, TX, USA
13. Department of Ophthalmology and Visual Sciences, University of Iowa Hospital and Clinics, Iowa City, IA, USA
14. Electrical and Computer Engineering, University of Iowa, Iowa City, IA, USA
15. Statistical Research and Data Science Center, Pfizer Inc, Groton, CT, USA
16. Department of Biomedical Informatics and Data Science, School of Medicine, Yale University, New Haven, CT, USA
17. Center for Biologics Evaluation and Research, Food and Drug Administration, Silver Spring, MD, USA
18. Epidemiology, Rutgers the State University of New Jersey, New Brunswick, NJ, USA

19. Statistical Editor, JAMA Network Open
20. Leonard Davis Institute of Health Economics, University of Pennsylvania, Philadelphia, PA, USA
21. Division of General Internal Medicine, Department of Medicine, Perelman School of Medicine, University of Pennsylvania, Philadelphia, PA, USA
22. Department of Health Data Science and Artificial Intelligence, McWilliams School of Biomedical Informatics, The University of Texas Health Science Center at Houston, Houston, TX, USA
23. Penn Medicine Center for Evidence-based Practice (CEP), Philadelphia, PA, USA
24. Penn Institute for Biomedical Informatics (IBI), Philadelphia, PA, US

*: Senior authors

**Corresponding authors:**  Yong Chen, Ph.D.
University of Pennsylvania
Blockley Hall 602, 423 Guardian Drive
Philadelphia, PA 19104
Office: 215-746-8155
E-mail: ychen123@upenn.edu

Xiaoqian Jiang, Ph.D.
UTHealth Houston
7000 Fannin St Suite 600
Houston, Texas 77030
E-mail: Xiaoqian.Jiang@uth.tmc.edu

ABSTRACT

Sample size calculations for power analysis are critical for clinical research and trial design, yet their complexity and reliance on statistical expertise create barriers for many researchers. We introduce PowerGPT, an AI-powered system integrating large language models (LLMs) with statistical engines to automate test selection and sample size estimation in trial design. In a randomized trial to evaluate its effectiveness, PowerGPT significantly improved task completion rates (99.3% vs. 88.9% for test selection, 99.3% vs. 77.8% for sample size calculation) and accuracy (94.1% vs. 55.4% in sample size estimation, $p < 0.001$), while reducing average completion time (4.0 vs. 9.3 minutes, $p < 0.001$). These gains were consistent across various statistical tests and benefited both statisticians and non-statisticians as well as bridging expertise gaps. Already under deployment across multiple institutions, PowerGPT represents a scalable AI-driven approach that enhances accessibility, efficiency, and accuracy in statistical power analysis for clinical research.

## INTRODUCTION

Clinical trials are an essential part of clinical research. Statistical power analysis is a cornerstone of study design in clinical research, enabling investigators to estimate the likelihood of correctly detecting a true effect[1–5]. Properly powered studies are essential to accurately estimate effect sizes, in addition, underpowered or overpowered trials could lead to incorrect results, wasted resources, and ethical concerns[6–9]. However, selecting appropriate statistical methods and accurately calculating sample sizes remain significant challenges, particularly for non-statisticians[10].

Access to statistical expertise is limited[11]. Even in well-resourced settings, limited statistician time can delay study design and analysis[12]. Existing software solutions for power analysis, such as G*Power[13], PASS[14], and R-based packages like pwr[15], have facilitated these calculations, but they often require users to have foundational knowledge in statistics or programming and be comfortable with command-line interfaces. More advanced platforms like nQuery[16] offer comprehensive tools for adaptive and group-sequential trials, while web-based resources like trialdesign.org[17] provide interactive support for specific study designs. However, these solutions remain largely menu-driven and require users to navigate complex parameter settings without interactive, real-time guidance. Furthermore, most traditional power analysis tools are designed for standard study designs and lack flexibility for non-standard scenarios, such as multi-arm trials, adaptive methodologies, or studies requiring real-time data integration for effect size refinement.

Recent advancements in artificial intelligence (AI), particularly large language models (LLMs) like OpenAI's GPT, offer promising solutions for automating complex tasks and provide great potential to inform regulatory decision-making[18–24]; however, their direct application to clinical trial power analysis has notable limitations. Errors in post-hoc power calculations and tests of normality have been observed when using general-purpose LLMs such as ChatGPT[25], underscoring the need for domain-specific adaptation and rigorous evaluation when applying LLMs in statistical workflows[26–29].

To address these challenges, we propose PowerGPT, a freely available AI-driven system designed to simplify access to complex statistical power analysis in clinical trial design. Unlike traditional pretrained-finetuned models, PowerGPT functions as an agent-based, end-to-end system, aiming to enable one-click power analysis by integrating LLMs with statistical software, computational engines, and external databases. It is designed to provide interactive guidance, natural language explanations, and adaptive recommendations to support study design.

Unlike most GPT-based statistical applications, which remain in experimental or small-scale deployment phases, PowerGPT has been piloted in multiple institutions and under deployment across NIH Clinical and Translational Science Award sites. Its integration into real-world research settings underscores the translational impact of AI in clinical research and demonstrates its potential to expand access to statistical expertise, accelerate study design, and enhance the rigor of clinical investigations. By combining AI-driven automation with domain-specific knowledge validation, PowerGPT represents a significant step toward reliable and interpretable AI deployment in biomedical research and clinical trial design.

We describe the design, use, and evaluation of PowerGPT, a freely available (https://power-gpt.net/) designed to perform sophisticated power analyses with minimal technical expertise. To evaluate PowerGPT's effectiveness, we conducted a balanced and stratified randomized trial comparing its performance against traditional sample-size estimate methods in University of Pennsylvania (UPenn) and University of Texas Health Science Center at Houston (UTHealth).

## RESULTS

### *Overall Agent-based architecture*

PowerGPT is an agent-based system that integrates user interactions, computational engines, databases, storage, and external tools to facilitate statistical power analysis. As illustrated in **Figure 1**, researchers interact with PowerGPT through a graphical user interface or a command-line console, submitting queries related to statistical test selection, sample size calculations, and result interpretation. These queries are processed by the internal API gateway, which directs requests to the appropriate system components. When freely available external data sources, such as statistical repositories or software libraries, are required, the external data exchange facilitates connections to web services and code repositories. These external tools, including third-party statistical libraries and APIs, provide specialized functions that expand PowerGPT's analytical capabilities. The system dynamically retrieves and integrates these resources as needed, ensuring seamless access to advanced computational methods.

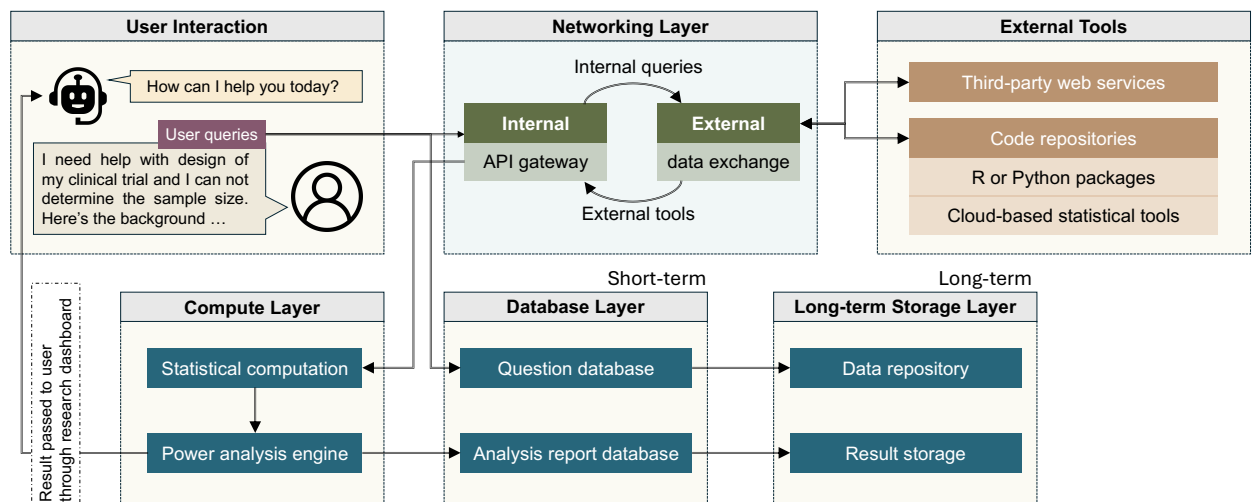

**Figure 1: PowerGPT architecture.** An integrated agent-based system combining user interfaces, computational engines, databases, storage, and external tools. The architecture leverages a dynamic function-calling mechanism to manage tasks such as data collection, analysis, and interpretation, ensuring scalability and efficiency for statistical power analysis.

The compute layer processes statistical analyses and power calculations, executing computations based on user queries. PowerGPT's underlying LLM assists in interpreting these queries, selecting appropriate statistical methods, and generating human-readable explanations of results. Once the calculations are complete, results are stored in the database layer, which consists of a short-term database that temporarily logs user queries and intermediate outputs for ongoing sessions, ensuring smooth and efficient real-time interactions. For longer-term needs, the storage layer retains structured datasets, computational results, and user-generated reports. This distinction allows users to revisit prior analyses without needing to re-run computations, supporting reproducibility and collaborative workflows. By tightly integrating these components, PowerGPT not only streamlines workflows but also translates complex statistical outputs into intuitive, user-friendly insights, making advanced power analysis accessible without requiring deep technical expertise.

PowerGPT supports a wide range of statistical tests commonly used in clinical and research settings. These include parametric tests such as one-sample t-tests, two-sample t-tests, paired t-tests, one-way ANOVA, and multi-way ANOVA. It also accommodates proportion-based analyses, including single-proportion and two-proportions z-tests, as well as Chi-square tests. For survival analysis, PowerGPT provides Cox proportional hazards models and log-rank tests. Additionally, it supports correlation tests, simple and multiple linear regression, and non-parametric methods such as the one-mean Wilcoxon test, Mann-Whitney test, paired Wilcoxon test, Kruskal-Wallis test, and non-parametric regression. Advanced methods, including repeated measures ANOVA, multi-level modeling, and generalized modeling approaches, are also integrated, ensuring flexibility across various study designs.

*Interactive Workflow for Statistical Analysis*
PowerGPT pairs a front-end of human-like interaction with a hidden back-end of advanced computational capabilities. The process begins with users describing their study's context or objectives in natural language, such as investigating differences in disease activity between groups or calculating sample sizes for a specific effect. PowerGPT prompts users for clarifications and additional details such as effect sizes, standard deviations, and desired power. This conversation mirrors how an investigator might discuss a project with a statistician in order to align investigator goals with the statistical task.

Once the study parameters are defined, PowerGPT identifies the appropriate statistical tests and provides detailed guidance on the required inputs. It returns statistical concepts in plain language. As the interaction progresses, PowerGPT dynamically adapts its recommendations based on user feedback or missing parameters. For instance, if users are unable to specify an effect size, PowerGPT suggests reasonable assumptions or requests alternative input data, iteratively tailoring the analysis to the user's needs.

When necessary statistical parameters are finalized, PowerGPT connects with external APIs and statistical engines to execute the necessary computations. Results are then returned to investigators in a colloquial and actionable format. If users wish to explore alternative scenarios—such as varying power levels or group sizes—PowerGPT revisits the analysis, recalculates the results, and provides updated recommendations. Throughout, the program accepts natural language prompts such as "Let's instead assume an 18% baseline incidence aiming to measure a 4% absolute risk reduction with treatment." Presented with such a statement, the program immediately re-calculates the required sample size.

A key feature of PowerGPT is its emphasis on transparency and user education. At each stage, it explains its choices and computations, helping users understand the statistical logic behind the recommendations. **Figure 2** provides an example of this workflow, illustrating how PowerGPT guides users from defining study objectives to delivering precise, actionable results.

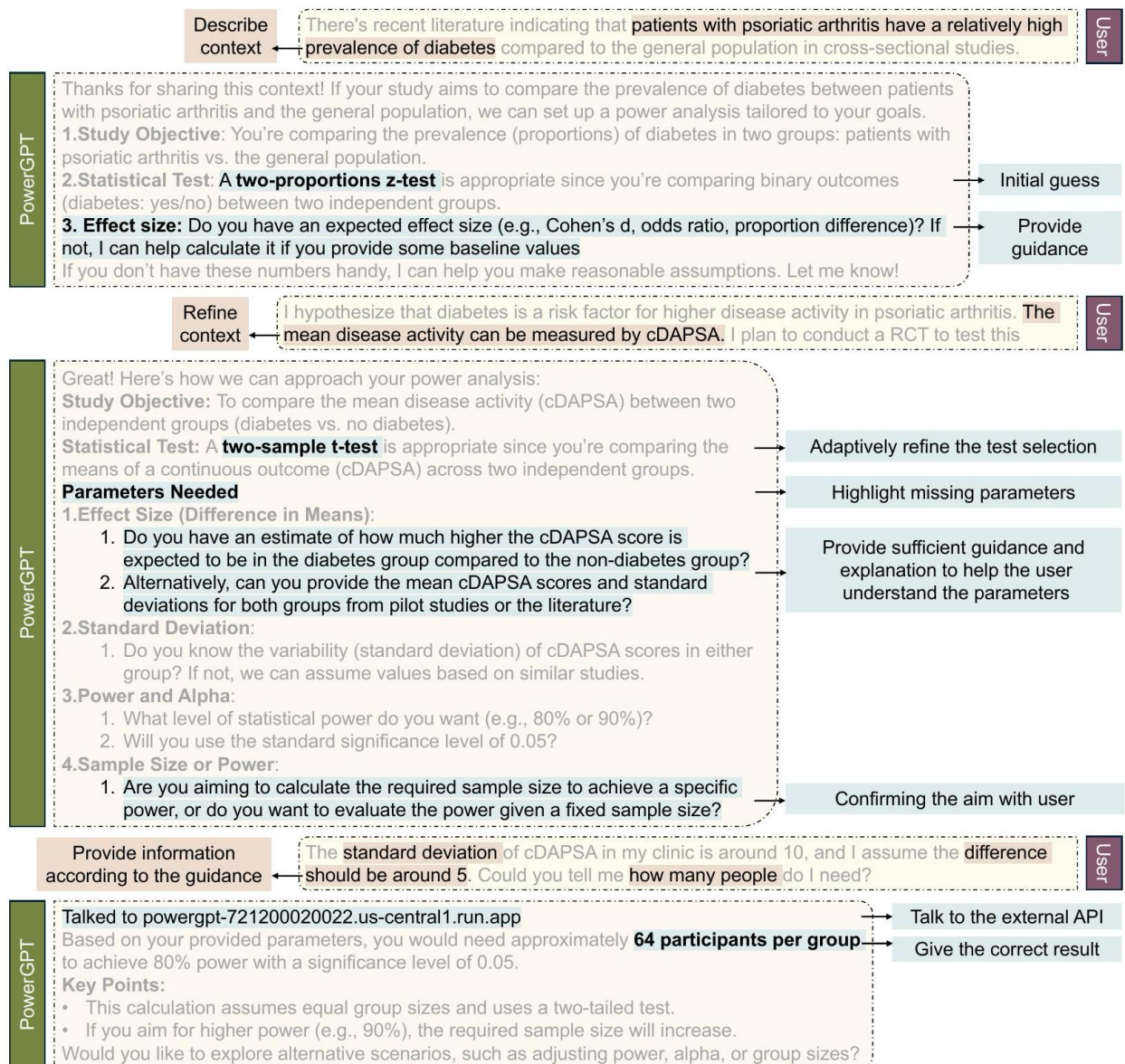

**Figure 2: Work process of PowerGPT.** The step-by-step workflow begins with understanding user objectives through natural language inputs and progresses to delivering actionable statistical results. The figure illustrates adaptive refinement of parameters, integration with external APIs, and presentation of outputs in a user-friendly format.

*Evaluation – Randomized Trial*

We evaluated the effectiveness of PowerGPT with a balanced and stratified randomized trial of 36 participants from UPenn and UTHealth. Participants represented a range of expertise, including statisticians (mathematics, statistics, and biostatistics) and non-statisticians (computer science, informatics, and biomedical science). To ensure balanced representation and statistical power, we employed a stratified randomization approach, assigning nine participants from each institution to the PowerGPT group and nine to the reference group.

Participants assigned to the PowerGPT group received access to the tool along with brief training on its use for statistical power analysis. They were instructed to rely solely on PowerGPT for test selection and sample size calculation. In contrast, those in the reference group used traditional methods, including textbooks, statistical software, and online search engines such as Google, but were not permitted to use any AI-powered tools. This design ensured that any observed differences in performance could be attributed specifically to PowerGPT rather than general AI assistance.

Both groups completed the same set of eight statistical test scenarios, covering widely used methods in clinical research, including one-sample t-test, two-sample t-test, paired t-test, one-way ANOVA, single-proportion z-test, two-proportions z-test, Cox proportional hazards model, and log-rank test. For each scenario, participants were required to identify the correct statistical test and determine the minimum sample size needed to achieve a predefined level of statistical power within three-hour time limit. The statistical scenarios used in the trial are provided in **Supplementary Materials A**.

The outcomes of interest were task completion rate and accuracy in two tasks (test selection and sample size calculation), as well as the time spent on each question. Task completion rate was defined as the proportion of assigned problems that were successfully completed within the time limit. To establish a reference standard for evaluating the accuracy of statistical test selection, we relied on a committee of three highly experienced statisticians (JT, YC, and HC). Each independently reviewed the results. For this study, all three statisticians reached unanimous agreement on the correct answers in every case, ensuring a robust and reliable benchmark. Both completion rate and accuracy were assessed separately for the two tasks (test selection and sample size calculation). Accuracy was calculated only for completed cases, meaning that incomplete responses were excluded from accuracy assessment. Additionally, since an incorrect test selection inevitably leads to an incorrect sample size calculation, any incorrect test selection automatically resulted in the corresponding sample size calculation being classified as incorrect. Time was measured as the total time taken to complete both test selection and sample size estimation for each scenario. For time analysis, only scenarios where both tasks were completed were included, ensuring that time measurements reflected valid attempts at both test selection and sample size calculation.

Comparisons between groups were conducted using two-proportion z-tests for completion rates and accuracy, while two-sample t-tests were used to assess differences in time spent on each question. We also analyzed different scenarios to determine whether PowerGPT's performance varied by scenario type. In addition to these overall comparisons, we performed a stratified analysis to evaluate whether PowerGPT's effectiveness differed by domain expertise, analyzing results separately for statisticians and non-statisticians. The detailed protocol and statistical analysis plan are presented in **Supplementary Materials B**.

*Performance in Test Selection and Sample Size Estimation*
The evaluation achieved 97.2% response rate, and a detailed breakdown of participant characteristics is presented in **Table 1**. The results underscore notable differences in performance between PowerGPT group and the reference group across critical metrics, including task completion rates, accuracy, and time spent on each question. These findings are visually summarized in **Figure 3**, which comprises four detailed panels: (a) overall task completion rates and accuracy for two tasks, (b) comparative analysis of overall time required for each question, (c) completion rate and accuracy distribution across individual scenarios, and (d) time required for each question across individual scenarios. Together, these panels provide a comprehensive depiction of the performance landscape, highlighting the strengths and limitations of each group. Detailed results are presented in **Supplementary Materials C**.

|  | **PowerGPT Group (N = 17)** | **Reference Group (N = 18)** |
|---|---|---|
| **Highest degree** |  |  |
| Bachelor's | 3 (17.6%) | 2 (11.1%) |
| Master's | 7 (41.2%) | 8 (44.4%) |
| PhD | 7 (41.2%) | 8 (44.4%) |
| **Domain of expertise** |  |  |
| Statistician | 10 (58.8%) | 9 (50.0%) |
| Non-statistician | 7 (41.2%) | 9 (50.0%) |
| **Year of profession experience** | 4.1 (2.8) | 3.9 (2.5) |
| **Familiarity with power analysis** |  |  |
| None | 2 (11.8%) | 2 (11.1%) |
| Beginner | 8 (47.1%) | 10 (55.6%) |
| Intermediate | 5 (29.4%) | 6 (33.3%) |
| Advanced | 2 (11.8%) | 0 (0.0%) |
| **Previous usage of other power analysis software** |  |  |
| Yes | 2 (11.8%) | 8 (44.4%) |
| No | 15 (88.2%) | 10 (55.6%) |
| **Previous usage of ChatGPT** |  |  |
| Yes | 15 (88.2%) | 15 (83.3%) |
| No | 2 (11.8%) | 3 (16.7%) |

**Table 1. Baseline characteristics of study participants.** Summary of participants' demographics, academic background, domain of expertise, years of professional experience, familiarity with power analysis, prior use of power analysis software, and previous experience with ChatGPT.

As shown in **Figure 3(a)**, PowerGPT significantly improved every performance metric (P<0.001). For test selection, the completion rate was 99.3% (95% CI: 95.4%-100.0%) in the PowerGPT group, while the reference group achieved 88.9% (82.3%-93.3%), with a significant difference of 10.4%, 4.3%-16.4%, p < 0.001). The accuracy of test selection was also higher in the PowerGPT group (95.6% [90.2%-98.2%]) compared to the reference group (83.6% [75.8%-89.3%]), showing a significant difference of 12.0% (3.9%-20.0%, p < 0.001). Similarly, for sample size calculation, the completion rate was 99.3% (95.4%-100.0%) in the PowerGPT group versus 77.8% (69.9%-84.1%) in the reference group, with a difference of 21.5%, (13.8%-29.1%, p < 0.001). Accuracy in sample size calculation showed a similar pattern, with PowerGPT achieving 94.1% (88.3%-97.2%), significantly outperforming the reference group's 55.4% (45.7%-64.7%), with a difference of 38.7% (27.9%-49.6%, p < 0.001). Regarding time efficiency, **Figure 3(b)** shows that participants using PowerGPT completed each question in an average of 4.0 minutes (95% CI: 3.3-4.7 min) and a median of 2.9 minutes (IQR: 1.4-5.3 min), whereas those in the reference group took 9.3 minutes (7.2-11.5 min) on average and a median of 6.9 minutes (3.7-12.2 min), resulting in a mean difference of 5.3 minutes (3.1-7.6 min, p < 0.001). These results demonstrate

PowerGPT's capability to enhance both accuracy and efficiency in statistical test selection and sample size calculation.

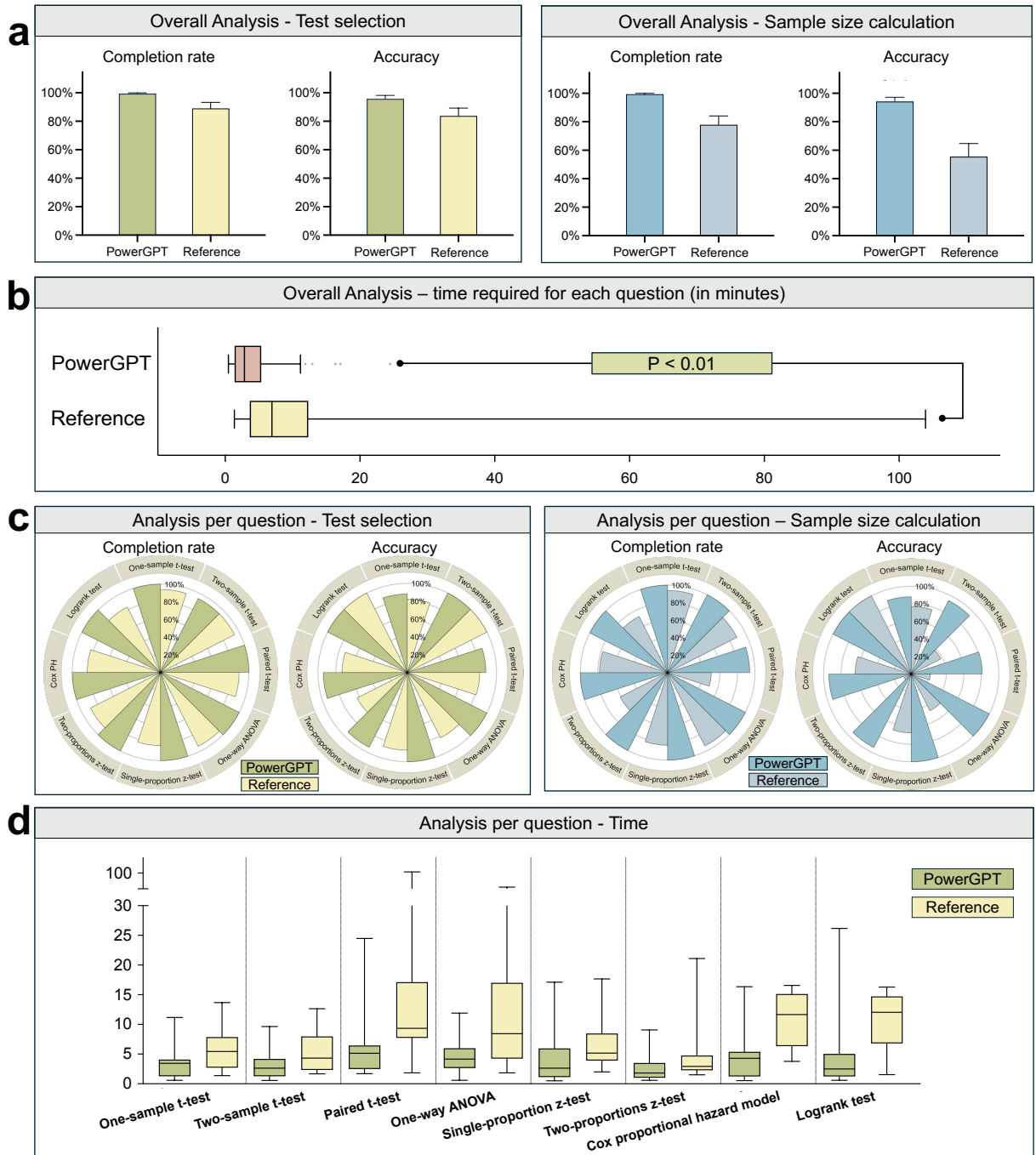

**Figure 3. Evaluation results of PowerGPT.** (a) Task completion rate and accuracy for test selection and sample size calculation in the PowerGPT and reference groups. PowerGPT users exhibited significantly higher completion rates and accuracy across both tasks. (b) Time required per question for both groups. PowerGPT substantially reduced the time required to complete statistical tasks compared to traditional methods. (c) Task completion rate and accuracy across different statistical tests. While PowerGPT maintained high performance across all tests, the reference group exhibited substantial variability, particularly in more complex tests. (d) Time required per question across different statistical tests. The reference group showed greater variability and longer completion times. PowerGPT provided more consistent and efficient performance across all scenarios.

*Task-Specific Performance Across Different Statistical Tests*

Completion, including test selection and in sample size calculation, by participants randomized to PowerGPT assistance were consistently high across all test scenarios, with each achieving 100% completion except for ANOVA (94.1%) in test selection and Cox proportional hazards model (94.1%). Accuracy varied slightly by test type but remained above 88.2% in test selection and 81.2% in sample size calculation, outperforming conventional approaches in all cases. In the reference group, accuracy and completion rates were more variable, with sample size calculation accuracy for the Cox proportional hazards model and log-rank test being particularly low at 22.2% and 9.1%, (**Figure 3c**).

The time required for each question varied considerably in reference participants, with some tests taking significantly longer than others. For instance, tasks involving the log-rank test and the Cox proportional hazards model had the longest completion times, with median times of 12.1 minutes (IQR: 7.7-14.3) and 11.7 minutes (IQR: 7.5-14.0). Unsurprisingly, simpler tests such as the two-sample t-test and two-proportion z-test had much shorter completion times at 3.8 minutes (IQR: 2.4-8.0) and 2.9 minutes (IQR: 2.4-4.7). PowerGPT not only reduced overall completion time but also reduced variability across different test scenarios, ensuring a more stable user experience for power analysis (**Figure 3d**).

*Impact of PowerGPT on Users with Different Statistical Expertise*

To further investigate PowerGPT's impact across different expertise levels, we performed a stratified analysis by grouping participants based on their domain of expertise. Participants with backgrounds in mathematics, statistics, and biostatistics were classified as statisticians, while those from computer science, informatics, and biomedical science were categorized under non-statisticians.

As shown in **Figure 4(a)**, non-statisticians without PowerGPT exhibited significantly lower performance in both test selection and sample size calculation compared to those with statistical expertise. Specifically, for test selection, statisticians without PowerGPT achieved a **100.0%** completion rate, whereas non-statisticians had a significantly lower completion rate of **77.8%** (95% CI: 66.2%-86.4%), with a notable difference of **22.2%** (11.2%-33.2%, p < 0.001). A similar trend was observed in sample size calculation, where statisticians achieved a **95.8%** (87.5%-98.9%) completion rate, while non-statisticians performed significantly worse, completing only **59.7%** (47.5%-70.9%) of tasks—a difference of **36.1%** (22.5%-49.7%, p < 0.001).

The introduction of PowerGPT significantly improved performance across both groups, effectively bridging the gap between expertise levels. In the PowerGPT group, non-statisticians achieved a 100.0% completion rate in both test selection and sample size calculation, closely matching the performance of statisticians, who had completion rates of 98.8% (92.3%-99.9%) for both tasks, with a non-significant difference of 1.2% (-2.4%-4.9%, p = 1). Beyond improving completion rates, PowerGPT also enhanced accuracy in both groups. Statisticians' accuracy increased to 94.9% (86.9%-98.4%) compared to statisticians without PowerGPT in test selection and 93.7% (85.2%-97.6%) in sample size calculation. Non-statisticians performed similarly, achieving 96.4% (86.6%-99.4%) in test selection, with a non-significant difference of 1.5% (-6.9%-9.8%, p = 1) compared to statisticians, and 94.6% (84.2%-98.6%) in sample size calculation, with a non-significant difference of 1.0% (-8.0%-9.9%, p = 1) relative to statisticians.

The improvement in the non-statisticians' group was particularly notable, as it demonstrates that PowerGPT not only elevated performance across all users but also narrowed differences between statisticians and non-statisticians. By offering structured guidance and statistical insights, PowerGPT enabled non-statisticians to achieve completion rate and accuracy levels comparable to those with formal statistical training, reinforcing its potential as an accessible and impactful tool for clinical research.

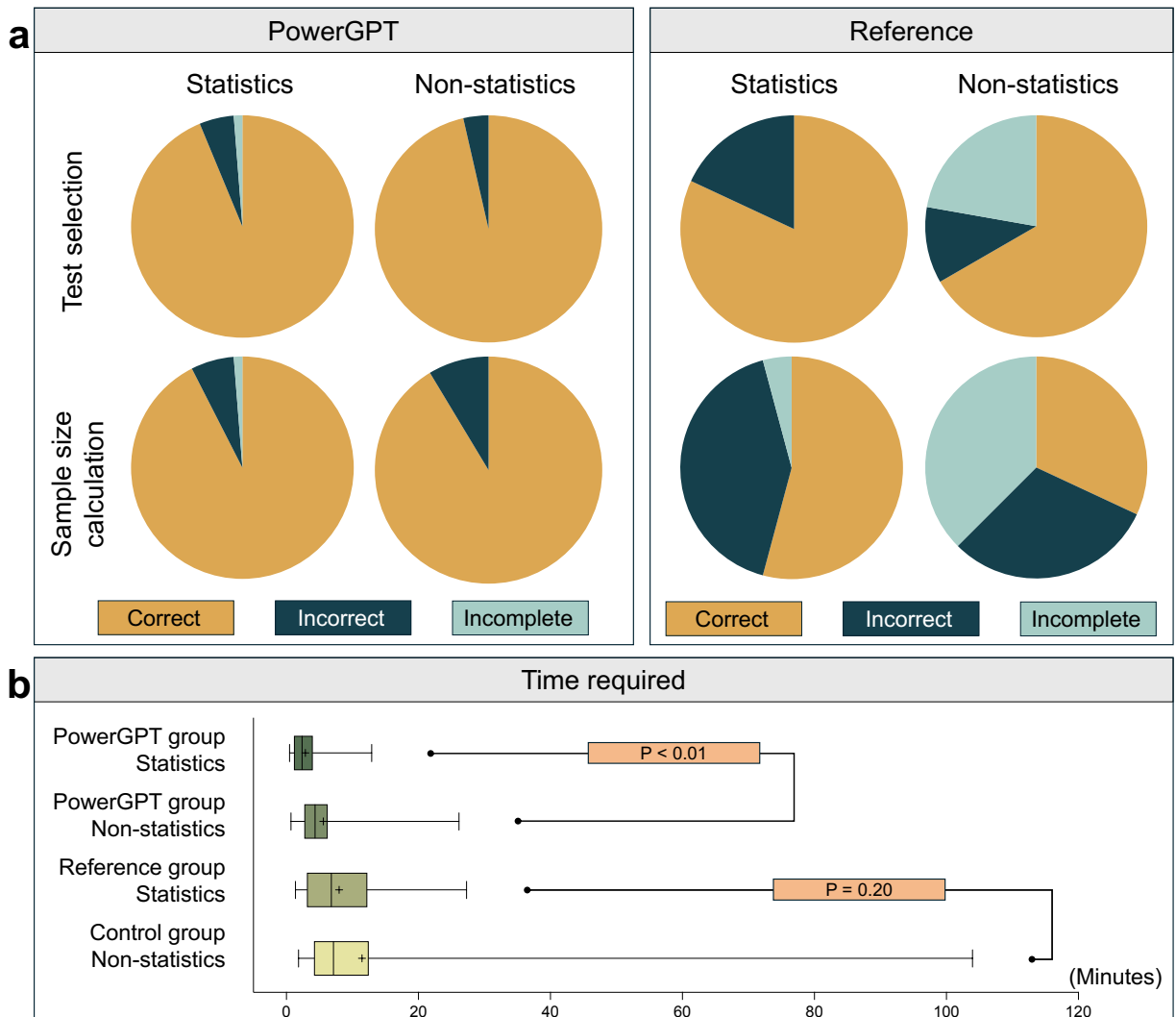

**Figure 4. Stratified analysis of PowerGPT's impact by domain of expertise.** (a) Task completion rate and accuracy stratified by expertise level. In the reference group, non-statisticians had significantly lower accuracy and higher incompletion rates compared to statisticians. PowerGPT improved performance across both groups and reduced the performance gap. (b) Time required per question stratified by expertise level. In the reference group, non-statisticians took significantly longer to complete tasks, with a long-tailed distribution indicating extreme delays for some participants.

**Figure 4(b)** illustrates the differences in time efficiency across subgroups. In the reference group, non-statisticians took significantly longer to complete tasks, with a median completion time of 7.1 minutes (IQR: 4.3-12.9) and an average completion time of 11.5 minutes (95% CI: 6.2-16.7) per question, compared to 8.0 minutes (6.6-9.4) on average and 6.8 minutes (3.2-12.1) in median for statisticians. Moreover, the non-statisticians' completion times in the reference group exhibited a particularly long tail, indicating that some participants required over 100 minutes to complete a task. Without PowerGPT, some researchers lacking statistical expertise therefore struggled with test selection and computational inefficiencies, potentially leading to inconsistencies in study design and delays in research workflows.

After using PowerGPT, the median completion times for both groups improved. Statisticians completed each task in 2.9 minutes (2.4-3.4 min) on average and 2.4 minutes (1.2-3.9 min) in median, while non-statisticians took slightly longer at 5.6 minutes (4.2-7.0 min) on average and 4.3 minutes (2.8-6.2 min) in median. Our results thus indicate that PowerGPT helped researchers with and without statistical expertise to overcome computational bottlenecks, thus making sample size calculation more efficient and accessible.

### *Deployment and Accessibility of PowerGPT*

PowerGPT is openly and freely accessible to researchers and institutions at [https://power-gpt.net/](https://power-gpt.net/), where users can explore its capabilities, review documentation, and integrate it into their research workflows. A dedicated feedback system enables users to provide insights, request new features, and contribute to the platform's continuous improvement.

PowerGPT has been piloted at multiple academic institutions, including the University of Pennsylvania, the University of Texas Health Science Center at Houston, Yale University, and others. It is also actively under deployment within the Clinical and Translational Science Award (CTSA) programs at UPenn and UTHealth, supporting clinical researchers in designing well-powered studies. Its integration into these institutions underscores its broad applicability and potential to enhance research efficiency by providing AI-driven statistical guidance tailored to different study designs. A list of the institutions that have piloted PowerGPT is available in **Supplementary Materials D**.

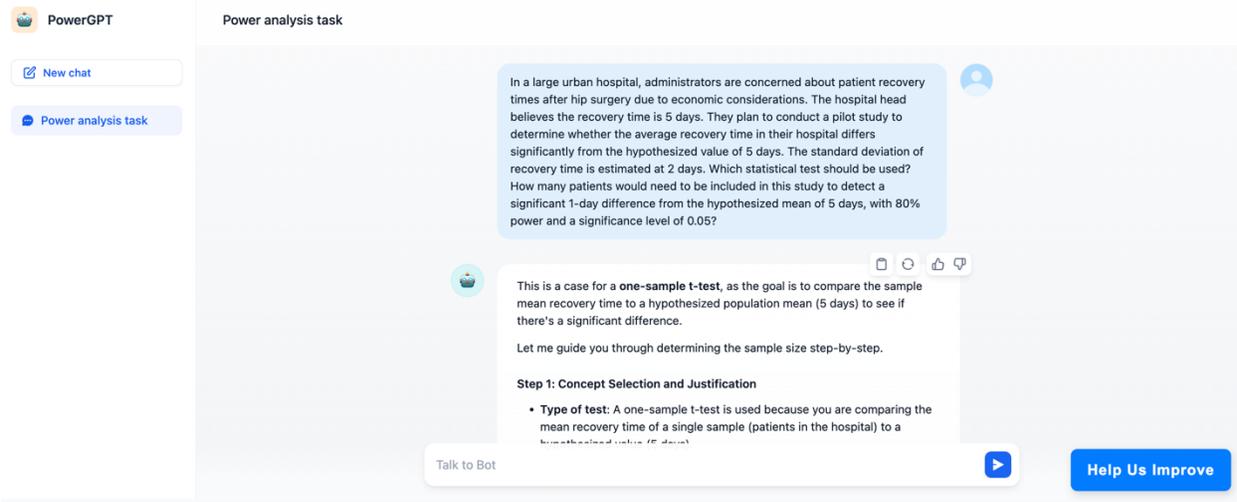

**Figure 5. PowerGPT Interactive Interface.** The interface allows researchers to interact with PowerGPT, select statistical tests, perform power calculations, and receive real-time guidance for study design.

## DISCUSSION

We introduce PowerGPT, a freely available agent-based system that integrates natural language processing with statistical computation to automate and enhance statistical power analysis. By providing an intuitive interface and adaptive recommendations, PowerGPT lowers the barriers to conducting sophisticated power calculations, making study design more accessible to researchers across various expertise levels. Beyond serving as a computational tool, PowerGPT represents a shift in how statistical guidance can be delivered—bridging the gap between traditional statistical software and the interactive, user-friendly capabilities of LLMs.

Our results demonstrate that PowerGPT significantly improves both accuracy and efficiency in sample size calculation. These benefits were particularly evident for non-statisticians but remained substantial for statisticians as well. By streamlining analytical workflows and minimizing errors, PowerGPT reduces the cognitive burden of statistical decision-making, allowing researchers to focus on study design rather than computational intricacies. Notably, it reduced performance gaps between statisticians and non-statisticians, a finding with important implications for increasing accessibility to rigorous study planning, particularly in settings where statistical expertise is limited.

This study has limitations. First, although this AI-enabled tool demonstrated high accuracy, the study focused on a predefined set of statistical tests, and its generalizability to more nuanced or domain-specific power analyses remains to be explored. Second, while PowerGPT demonstrated efficiency gains, further investigation is warranted to assess long-term usability, potential biases, and user trust in AI-assisted statistical decision-making. Third, PowerGPT is designed to facilitate the front end of research. Final analyses of data often require decisions about how to adjust for confounding or other considerations and how to report results in ways that are not misleading. These elements of the art and science of statistical reasoning continue to benefit from human expertise.

A key future direction for PowerGPT is enhancing error tolerance and validity checking to improve trustworthiness and reliability. Future iterations will focus on implementing mechanisms that identify invalid queries, verify input coherence, and provide step-by-step guidance supported by literature-backed best practices. These features will ensure that users receive context-aware recommendations, reducing the risk of misinterpretation and improving confidence in AI-assisted statistical reasoning. Additionally, PowerGPT is expected to evolve to include advanced statistical tests and to improve reasoning capabilities, contextual understanding, and real-time data integration to further refine its adaptability and effectiveness in clinical research applications.

In conclusion, this study provides empirical evidence that PowerGPT enhances the accuracy, efficiency, and accessibility of statistical power analysis. By integrating AI-driven tools into research workflows, clinical investigators can make more informed methodological choices, ultimately strengthening the quality and reproducibility of biomedical studies.

## METHODS

### *Standardized Statistical Methods Implementation*

To ensure PowerGPT's versatility and robustness, the preprocessing phase involved integrating widely recognized power analysis functions covering a broad spectrum of statistical tests. These range from commonly used methods, such as the two-sample t-test for comparing means, to more specialized tests, like the log-rank test, which is tailored for survival analysis in medical research and clinical trials. This comprehensive integration enhances the model's applicability across different research scenarios, making it suitable for a wide variety of study designs.

A detailed example of this preprocessing workflow can be seen in the integration of the log-rank test. This process begins with the careful selection of the appropriate statistical package and function. Specifically, we utilized the 'ssizeCT.default' function from the 'powerSurvEpi' package (version 0.1.3), a function grounded in Freedman's seminal method for survival analysis[19,30]. Following this, the necessary inputs were prepared for translation into a machine-readable format. An R document was created detailing the function's descriptions, arguments, and the associated R implementation. This step ensured that all relevant details were captured to facilitate seamless integration.

An auto-translator was employed to convert the prepared materials into API-recognizable code to make the R-based function accessible to the system. This translation process allowed the function to be dynamically called and executed by PowerGPT during its operations, enabling it to perform accurate and efficient power calculations. By streamlining these preprocessing steps, PowerGPT achieves a balance of precision and adaptability, equipping it to handle a wide range of statistical needs in research settings.

### *Prompt Engineering and Python-R Wrapper Integration*

The integration of external statistical functionalities in PowerGPT relies on a combination of robust prompt engineering and a Python-R API wrapper. These components ensure seamless communication between ChatGPT's assistant function calls, external APIs, and R-based statistical functions. Prompt engineering forms the foundation for guiding ChatGPT in constructing accurate function calls to external APIs. Clear API specifications with well-defined descriptions and parameters enable GPT to interact with services handling statistical tests such as Two-Sample T-Tests and Log-Rank Tests.

For instance, when handling a Two-Sample T-Test request, the system follows a structured workflow. First, ChatGPT captures the intent of the user query, such as "Calculate sample size using a Two-Sample T-Test." Based on the query, relevant inputs like delta (difference in means), sd (standard deviation), and power are mapped to the API schema. The chatbot then constructs an appropriate API request in JSON format, such as:

```
{
    "delta": 1.5,
    "sd": 0.5,
    "power": 0.8
}
```

These values are sent to the corresponding API for statistical computation. The system abstracts the complexity for users, allowing them to focus on statistical analysis rather than the specifics of API integration.

The Python-R API wrapper plays a critical role in enabling the system to access R's specialized statistical libraries while maintaining a scalable Python-based web service. Each statistical function, such as Two-Sample T-Tests or Log-Rank Tests, is exposed through distinct API endpoints defined using OpenAPI 3.1. For example:
- /api/v1/two_sample_t_test handles sample size calculations for Two-Sample T-Tests.
- /api/v1/log_rank_test performs survival analysis calculations.

When a valid API request is received, the Python backend triggers the corresponding R function via rpy 2 , such as power.t.test (). The results, typically returned in native R structures, are converted to JSON format for integration into the system's front end. For example, an API request for a Two-Sample T-Test might return:

```
{
    "sample_size":                                                      102
}
```

This approach provides a user-friendly output, allowing researchers to immediately interpret the results.

***Cloud-Based Infrastructure and Automated Computation***

PowerGPT is built on a state-of-the-art Large Language Model (LLM) integrated with a cloud-native infrastructure to optimize statistical power analysis. The platform orchestrates user interactions and computational tools to streamline complex statistical workflows. External tools are integrated via function calls, allowing PowerGPT to automate statistical computations using freely available R or Python-based libraries. This approach supports various statistical functions, including tests like Chi-Squared Tests, Paired T-Tests, and Log-Rank Tests, making the platform suitable for various research scenarios.

The system also employs preprocessing and translator modules to map user queries to appropriate statistical functions. For example, parameters for a T-Test (e.g., mean difference, standard deviation, and power requirements) are automatically converted into JSON format for compatibility with R statistical packages. Robust error-checking mechanisms validate inputs such as effect sizes and probabilities to ensure meaningful results.

User-friendly prompts and dynamic validation enhance the natural interaction between researchers and PowerGPT. These features guide users through the analysis process, reducing barriers for those without advanced statistical knowledge. The workflow involves task identification and mapping, API calls to external modules, and result compilation. For example, when a user requests a Two-Sample T-Test, the system identifies the task, executes the appropriate R function, and presents the results in clear language.

By combining well-structured prompts, a robust Python-R integration, and a comprehensive statistical library, PowerGPT provides a scalable and user-friendly solution for clinical researchers and data analysts seeking to optimize trial designs and statistical analyses.

**PRIVACY AND DATA SECURITY**
Ensuring the privacy and security of sensitive research data is a critical priority for PowerGPT. The platform is under deployment on UTHealth Google Cloud Run, which is HIPAA-compliant, providing a secure and scalable environment for statistical computations. Additionally, PowerGPT leverages the UTHealth Enterprise OpenAI Service, a HIPAA-compliant workspace that guarantees OpenAI does not retain user data or use it for model training. This setup fully aligns with UTHealth's stringent security policies, ensuring data confidentiality, integrity, and compliance with industry standards.

**CODE AVAILABILITY**
PowerGPT backend server is currently hosted on Google Cloud, providing robust scalability and performance for concurrent processing. However, the system is fully containerized and can be easily deployed on other Docker platforms (AWS Elastic Beanstalk or Azure Web App). The full source code for PowerGPT, including its infrastructure, API definitions, and integration with R statistical functions, is openly available on GitHub at: https://github.com/x1jiang/pwgpt

*System Scalability and Concurrency*
PowerGPT is built on OpenAI's ChatGPT action API and Google Cloud Run, combining AI-driven computation with cloud-based scalability. This architecture enables seamless performance under heavy loads while dynamically scaling from 0 to 200+ instances in real-time, ensuring availability even during unpredictable traffic surges. Our fully containerized system allows for rapid instance spin-up, optimizing resource utilization. Startup Probe configurations enhance resilience by continuously monitoring service readiness via a TCP probe on port 5000. If an issue is detected, the failure threshold (1) ensures immediate instance replacement, maintaining continuity at 120-second intervals with no startup delay. Unlike prototype systems that struggle with concurrency, PowerGPT is engineered for industrial-scale deployment. It efficiently manages complex statistical workflows, clinical trial setups, and high-volume analytical tasks—offering unmatched reliability, scalability, and responsiveness

A standout feature in our implementation is the Startup Probe configuration, tailored for maximum resilience and rapid startup. It is defined with the following parameters:
- TCP probe on port 5000: Ensures the service is ready only when it can healthily accept traffic.
- Interval: Every 120 seconds.
- Initial delay: 0 seconds, ensuring the system becomes functional as soon as possible without excessive pre-warm time.
- Timeout: 120 seconds.
- Failure threshold: A robust setting of 1 , meaning that if the check fails even once, the instance is immediately considered cold and replaced.

This full container-based setup, combined with advanced startup probe mechanics, differentiates our platform from traditional prototype designs that typically only support a few concurrent instances and generally serve as proof-of-concept systems. Unlike those smaller-scale systems, our platform is designed for industrial-scale deployment, ensuring that massive concurrent requests, complex function call sequences, and external API queries are handled efficiently without slowing down or stalling.

This robustness and scalability allow PowerGPT to support various procedures, analyses, and interactions simultaneously, making our system suitable for real-world problems and surpassing the limitations of typical academic or test-prototype designs. Whether processing multiple concurrent clinical trial setups or large-scale statistical power determinations, our system ensures consistency and speed at any scale.